\documentclass[letterpaper]{article} 
\usepackage{aaai24}  
\usepackage{times}  
\usepackage{helvet}  
\usepackage{courier}  
\usepackage[hyphens]{url}  
\usepackage{graphicx} 
\usepackage{natbib}  
\usepackage{caption} 
\usepackage{algorithm}
\usepackage{algorithmic}

\usepackage{newfloat}
\usepackage{listings}
\DeclareCaptionStyle{ruled}{labelfont=normalfont,labelsep=colon,strut=off} 
\floatstyle{ruled}
\newfloat{listing}{tb}{lst}{}
\floatname{listing}{Listing}
\title{AACP: Aesthetics Assessment of Children’s Paintings Based on
	\\ Self-Supervised Learning}
\author {
    Shiqi Jiang\textsuperscript{\rm 1, 3},
    Ning Li\textsuperscript{\rm 1},
    Chen Shi\textsuperscript{\rm 1},
    Liping Guo\textsuperscript{\rm 2},
    Changbo Wang\textsuperscript{\rm 1},
    Chenhui Li\textsuperscript{\rm 1}\thanks{Corresponding author}
}
\affiliations {
    \textsuperscript{\rm 1}School of Computer Science and Technology, East China Normal University\\
    \textsuperscript{\rm 2}Faculty of Education, East China Normal University\\
    \textsuperscript{\rm 3}Shanghai Institute of AI for Education, East China Normal University\\
    
    \{52265901032, 51215901058, 52205901019\}@stu.ecnu.edu.cn \\ lpguo@pie.ecnu.edu.cn, \{cbwang, chli\}@cs.ecnu.edu.cn
}

\usepackage{bibentry}

\begin{document}

\maketitle

\begin{abstract}
The \textbf{A}esthetics \textbf{A}ssessment of \textbf{C}hildren's \textbf{P}aintings (AACP) is an important branch of the image aesthetics assessment (IAA), playing a significant role in children's education. This task presents unique challenges, such as limited available data and the requirement for evaluation metrics from multiple perspectives. However, previous approaches have relied on training large datasets and subsequently providing an aesthetics score to the image, which is not applicable to AACP. To solve this problem, we construct an aesthetics assessment dataset of children's paintings and a model based on self-supervised learning. 1) We build a novel dataset composed of two parts: the first part contains more than 20k unlabeled images of children's paintings; the second part contains 1.2k images of children's paintings, and each image contains eight attributes labeled by multiple design experts. 2) We design a pipeline that includes a feature extraction module, perception modules and a disentangled evaluation module. 3) We conduct both qualitative and quantitative experiments to compare our model's performance with five other methods using the AACP dataset. Our experiments reveal that our method can accurately capture aesthetic features and achieve state-of-the-art performance.
\end{abstract}

\section{Introduction}
Aesthetics education plays a crucial role in the holistic development of children as it fosters the development of aesthetic skills, stimulates creativity, improves cultural literacy, and enhances social skills~\cite{SRAES}. Children's painting is a way for children to express their emotions, feelings and understanding of things. It is a reflection of their cognitive and emotional development, as well as their ability to perceive and interpret the world around them~\cite{Robson2012ObservingYC,Chang2005ChildrensDS}. Aesthetics assessment of children's paintings is a crucial component of aesthetics education, as it provides a way to measure and perceive the aesthetic qualities of children's paintings from multiple perspectives. By analyzing quantitative attributes, such as composition and color,  researchers can gain a more comprehensive understanding of the cognitive processes and artistic abilities of children.

\begin{figure}[H]
	\centering
	\includegraphics[width=1\linewidth]{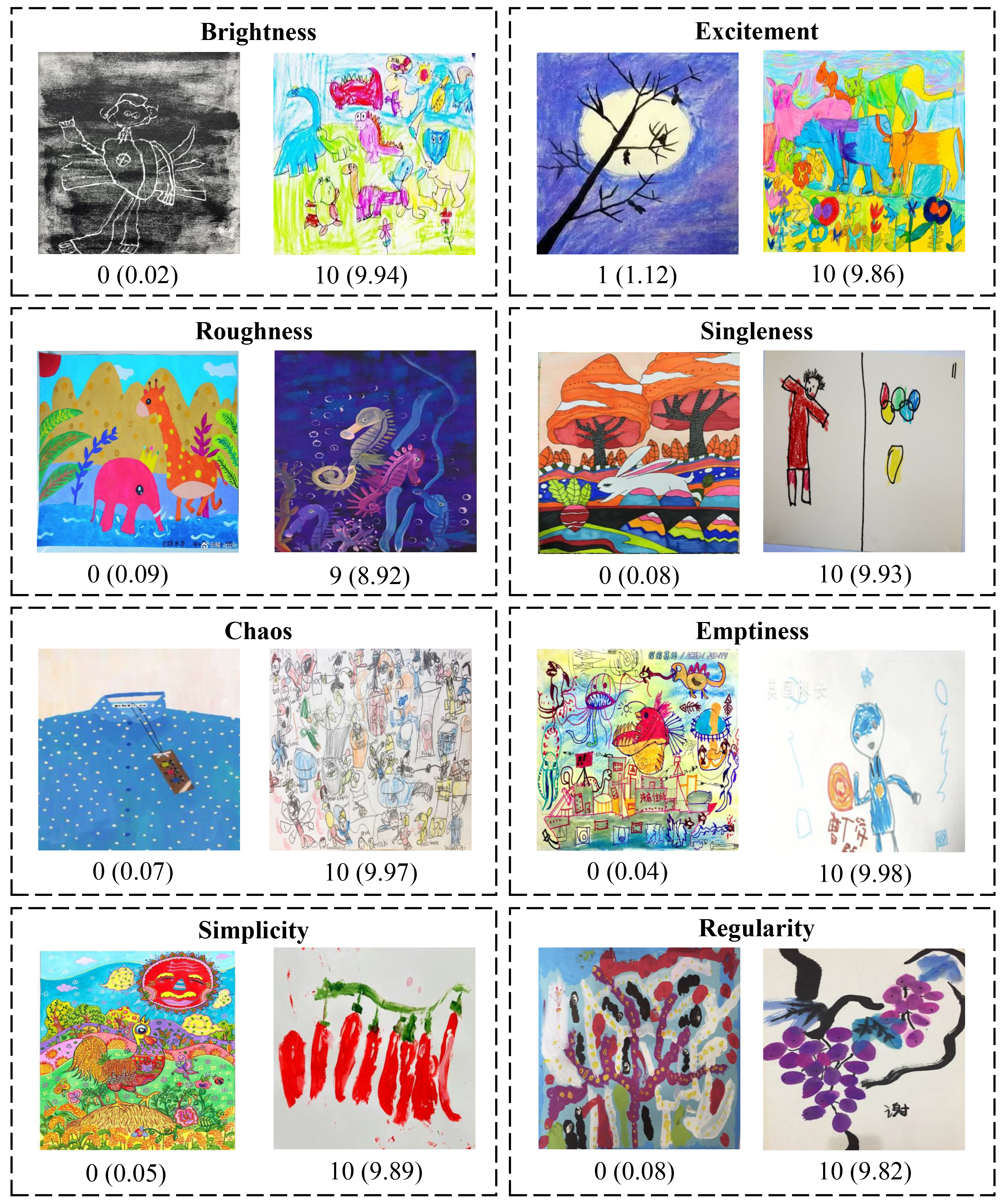}
	\caption{Examples of images and annotations in the proposed dataset with the ground truth (and predicted) scores at the bottom. We assess the aesthetics of children's paintings from 8 attributes.}
	\label{introduction1}
\end{figure}

Traditionally, the aesthetics assessment of children's paintings was performed by experts in the field of art or design,  who would evaluate the content, colors, and other aspects to infer the meaning or message behind the painting~\cite{Sali2014AnAO}. However, such methods are inherently subjective and may be influenced by personal bias or preconceived notions. In addition, their assessment metric is relatively single and potentially insufficient to capture the full range of potential meanings in children's paintings.

In recent years, researchers have explored the application of deep learning algorithms for assessing the aesthetic quality of paintings~\cite{DBLP:journals/tip/TalebiM18,DBLP:conf/cvpr/SheLY021}. This approach potentially simplifies the assessment process and offers a more objective and consistent way to assess the aesthetic value of paintings. Consequently, we have incorporated deep learning methods into the aesthetics assessment of children's paintings, aiming to improve the accuracy and objectivity of our assessment. However, this approach has also faced two challenges.

First, existing IAA datasets such as AVA~\cite{DBLP:conf/cvpr/MurrayMP12} and AADB~\cite{DBLP:conf/eccv/KongSLMF16} are not well-suited for the aesthetics assessment of children's paintings for two main reasons. 1) The majority of images in these datasets are natural images or photographs, which significantly differ from children's paintings. As shown in Figure~\ref{introduction1}, children's paintings often have unique characteristics, such as abstract and personalized styles, as well as simple and vivid compositions, which may not be effectively represented in existing datasets. 2) The single scores in these datasets inadequately represent children's paintings' broad aesthetic range and potential meanings that often express their emotions and cognitive development.

Second, previous IAA approaches used open datasets and direct image-to-score mapping~\cite{DBLP:journals/tip/TalebiM18}, achieving state-of-the-art performance on natural images using various techniques, such as multi-scale representation~\cite{DBLP:conf/iccv/KeWWMY21}, graph convolution networks (GNNs)~\cite{DBLP:conf/cvpr/SheLY021}, and other methods. However, such approaches may not be well suited for the aesthetics assessment of children's paintings, as they fail to adequately capture the intrinsic meanings and unique characteristics of these images, and it is insufficient to rely on a single score to evaluate the aesthetic quality of children's paintings.

To overcome existing challenges in assessing the aesthetics of children's paintings, we have made two key contributions. First, we have constructed the first AACP-specific dataset, as shown in Figure~\ref{introduction1}. Our dataset, the first of its kind specifically designed for AACP, provides a valuable resource for studying children's painting aesthetics and art-creation psychology. This dataset enables researchers to gain a more comprehensive understanding of these aesthetics, facilitating  aesthetics education and children's artistic development. Second, we propose a network architecture for AACP that comprises a masked encoder, two perception modules and a disentangled evaluation module. This architecture offers an accurate method for assessing children's painting aesthetics, further prompting children's artistic growth and aesthetics education.

In summary, our main contributions include:

$\bullet$ We have constructed a novel and multi-attribute dataset for the aesthetics assessment of children's paintings, specifically tailored to support children's education.

$\bullet$ We propose an effective model based on self-supervised learning to extract aesthetic features for AACP, eliminating the need for extensive manual annotations.

$\bullet$ Our approach demonstrates outstanding results on the AACP dataset through extensive experimentation, surpassing the performance of other methods for assessing the aesthetic qualities of images. 

\section{Related Work}
\subsection{Self-Supervised Learning}
Self-supervised learning, particularly contrastive learning and masked auto-encoding, is frequently utilized in computer vision. Contrastive learning aims to identify common features among positive samples while distinguishing differences between negative samples. One notable method is SimCLR~\cite{DBLP:conf/icml/ChenK0H20}, which employs a simple framework and has demonstrated superior performance over previous methods. Masked auto-encoding, on the other hand, encourages the model to learn local features of the data by partially masking the input data. Both MAE~\cite{DBLP:conf/cvpr/HeCXLDG22} and ConvMAE~\cite{DBLP:journals/corr/abs-2205-03892} have demonstrated success in obtaining excellent representations using masked auto-encoding. Furthermore, self-supervised learning can also be implemented through different masked views, as shown in works like data2vec~\cite{DBLP:conf/icml/BaevskiHXBGA22} and MaskFeat~\cite{DBLP:conf/cvpr/00050XWYF22}. These pre-trained models obtained through self-supervised learning can be effectively employed in downstream tasks such as classification and segmentation~\cite{DBLP:conf/iclr/Bao0PW22,DBLP:conf/iccv/CaronTMJMBJ21}.

In our study, we employ self-supervised learning to primarily reduce the need for labeled data and thus mitigate the cost of manual labeling. We further optimize the training and inference speed by simplifying the ConvMAE structure, facilitating the efficient assessment of results.
\begin{figure*}
	\centering
	\includegraphics[width=1\linewidth]{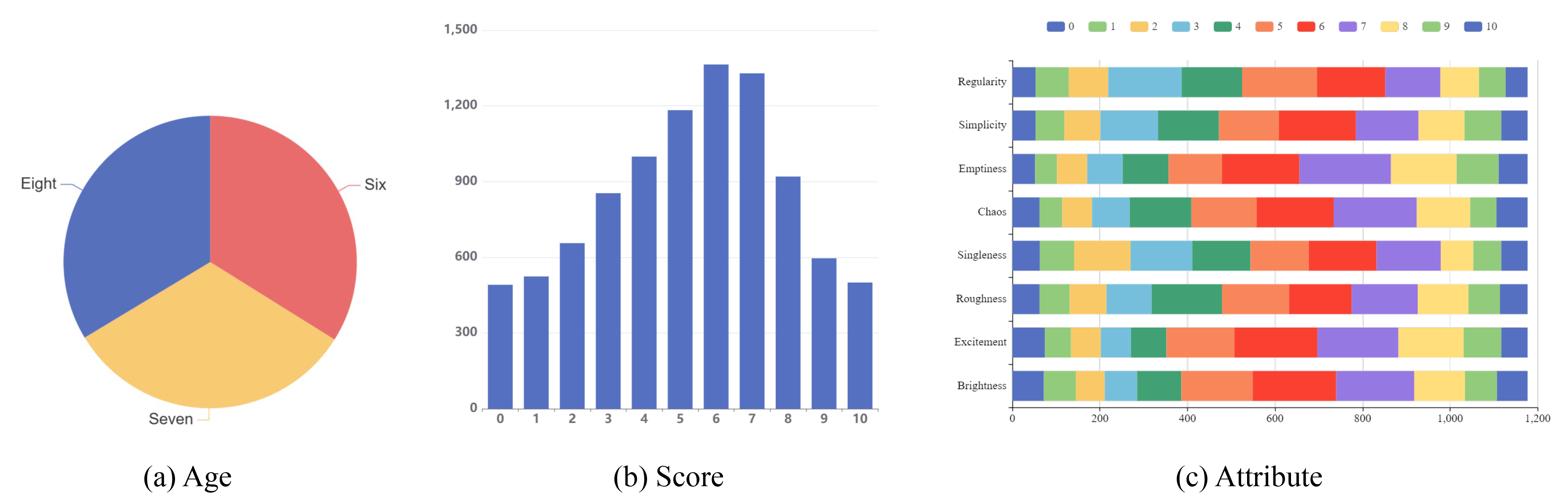}
	\caption{The age, score and attribute distribution of AACP dataset.}
	\label{dataset}
\end{figure*}
\subsection{IAA Model}
Early IAA models map aesthetics features from images to scores~\cite{DBLP:conf/mm/LuLJYW14,DBLP:conf/iccv/LuLSMW15,DBLP:conf/cvpr/MaiJL16}. Following this, A\_LAMP~\cite{DBLP:conf/cvpr/MaLC17} predicts aesthetics scores based on layout and patch. NIMA~\cite{DBLP:journals/tip/TalebiM18} produces a distribution of aesthetics ratings closely matching human ratings. MP$_{ada}$\cite{DBLP:conf/mm/ShengDMMHH18} uses multi-patch and attention mechanisms to improve learning efficiency when only aesthetics labels are available. MLSP\cite{DBLP:conf/cvpr/HosuGS19} and MUSIQ~\cite{DBLP:conf/iccv/KeWWMY21} propose a structure that maintains original image resolution as input, handling images of arbitrary size without information loss caused by cropping or scaling. UGIAA~\cite{DBLP:journals/tmm/LvFNDJZXX23} uses reinforcement learning and HGCN~\cite{DBLP:conf/cvpr/SheLY021} employs graph networks, while PIAA~\cite{DBLP:journals/tcyb/ZhuLWZDS22} leverages meta-learning. These diverse methods have advanced image aesthetics assessment from different perspectives.

However, the significance of spatial information in elements and the insufficiency of relying on labels from image aesthetics datasets for understanding children's paintings have been overlooked. Thus, we propose to explore spatial and channel features to improve AACP.

\subsection{IAA Dataset}
In recent years, several datasets have been constructed for image aesthetics assessment. AVA~\cite{DBLP:conf/cvpr/MurrayMP12} is a large-scale dataset that contains over 250k images with aesthetics, semantic, and style labels, and has been widely used in aesthetics assessment. After this, AADB~\cite{DBLP:conf/eccv/KongSLMF16} includes annotations for eight aesthetic factors, but the labels are not detailed enough to capture the diversity of aesthetics. PCCD~\cite{DBLP:conf/iccv/ChangLC17} first used linguistic comments with multiple aesthetic factors, which has comprehensive annotation but fewer data. Art500k~\cite{DBLP:conf/mm/MaoCS17} and OmniArt~\cite{DBLP:journals/corr/abs-1708-00684} are large-scale art painting datasets, but the data are collected from paintings in museums rather than photographs. SemArt~\cite{DBLP:conf/eccv/GarciaV18} is a dataset for semantic art understanding that includes attributes and textual art comments for each image, but is not specific to children's paintings. While these datasets have advanced the field of image aesthetics assessment in different areas, they are not suitable for the aesthetics assessment of children's paintings.

\section{AACP Dataset}
\subsection{Painting Collection}
To collect high-quality children's paintings, we first invited hundreds of children to participate in painting, whose age distribution is shown in Figure~\ref{dataset} (a). We did not constrain the content or duration of painting, in order to ensure that the collected works are complete and diverse. The collected paintings include works created using various tools such as oil pastels, watercolor pens, crayons, and brushes, as well as composite materials made using collages. 

Secondly, we invited ten experts in the design field to screen and annotate the paintings. The quantity of the manually labeled image was approximately 1.2k. To ensure the quality of the data, we set strict standards for the selection and labeling process. The labeled data was then used for training and testing our model. We believe that our approach of collecting and annotating children's paintings can be used as a reference for future studies in this area.

\subsection{Painting Annotation}
Each image is scored on a scale of 0 to 10 for each attribute, with a higher score indicating a closer alignment with that metric. Based on design theory and composition principles, experts have divided the assessment of children's paintings into four aspects: color, texture, composition, and conception. To make the annotation process easier, we use \textit{brightness} and \textit{excitement} to describe \textbf{color}, \textit{roughness} and \textit{singleness} to represent \textbf{texture}, \textit{chaos}, \textit{emptiness} and \textit{simplicity} to analyze \textbf{composition}, and \textit{regularity} to represent \textbf{conception}. In total, we use eight attributes to assess children's paintings. The distribution of scores and attributes in our dataset is shown in Figure~\ref{dataset} (b) and (c), respectively. We have balanced the amount of data for each score and attribute to avoid long-tailed distributions.
Please refer to the \textbf{supplementary materials} for an explanation of this annotation method. 

Specifically, each image is evaluated by multiple experts, and each attribute receives multiple scores. The final annotation value for each attribute is the average of these scores. By using this method, we can reduce the influence of subjective factors in the evaluation process.

\subsection{Dataset Expansion}
To overcome manual labeling constraints, we augment our children's painting dataset using generative models. While high-performance generators like StyleGAN2~\cite{DBLP:conf/cvpr/KarrasLAHLA20} and DDPM~\cite{DBLP:conf/nips/HoJA20} create realistic images, their lack of control limits their suitability. Instead, we focus on semantic-based image generation methods, such as DALL-E~\cite{DBLP:journals/corr/abs-2204-06125}, Imagen~\cite{DBLP:journals/corr/abs-2205-11487}, and Stable Diffusion~\cite{DBLP:conf/cvpr/RombachBLEO22} for controlled and high-quality generation. In our work, we specifically use DALL-E to generate children's painting images based on keyword combinations.

We carefully choose semantic keywords related to children's painting attributes—such as tone, pattern, and composition—to maximize image diversity and realism. Employing these keywords with DALL-E, we generate a diverse dataset of approximately 20,000 images.

We employ this approach for two primary reasons. Firstly, the Dall-E model produces diverse, compliant children's paintings in large quantities, with prompts ensuring diversity. Secondly, the Fréchet Inception Distance (FID) metric~\cite{DBLP:conf/nips/HeuselRUNH17} scored 7.85, suggesting high similarity between generated and real paintings. This allows us to pre-train our model using generated images.

\subsection{Annotation Explanation}
\textbf{Color}~~We assess the color in children's paintings based on brightness and excitement. Brightness denotes the lightness or darkness of a color, with lighter colors having high brightness and darker colors having low brightness. Excitement pertains to color saturation and vibrancy; bright, saturated colors evoke excitement, whereas darker, unsaturated hues suggest calmness. Thus, we assess a painting's color using both brightness and excitement.

\textbf{Texture}~~The texture in children's paintings is analyzed based on factors such as roughness versus smoothness, and singularity versus complexity. Rough textures may indicate a child's anxiety or unrefined brush control, while smooth textures could suggest relaxation or confidence. A painting with a singular texture may point towards a lack of interest or creativity, while complex textures might reflect high levels of excitement or creativity.

\textbf{Composition}~~The composition of children's paintings, which includes element arrangement and tone relationships, conveys themes and aesthetics. We assess composition by examining the degrees of chaos, emptiness, and simplicity. Chaos might represent a child's emotional unrest or an unclear idea. Emptiness could signify introversion, a creativity deficit, or a lack of motivation. Simplicity, implying the level of detail, may suggest a limited imagination if the composition lacks complexity.

\textbf{Conception}~~We assess the concept of a child's painting by examining the level of regularity present in the work. A regular concept can be interpreted as an indication of the child's self-confidence in their abilities or trust in their judgment during the creative process, or it may reflect the child's imaginative thinking at the time of drawing. Conversely, an irregular concept may represent the child's curiosity or creativity, or it could be a sign of strong critical thinking skills demonstrated during the drawing process.
\begin{figure*}[h]
	\centering
	\includegraphics[width=1\linewidth]{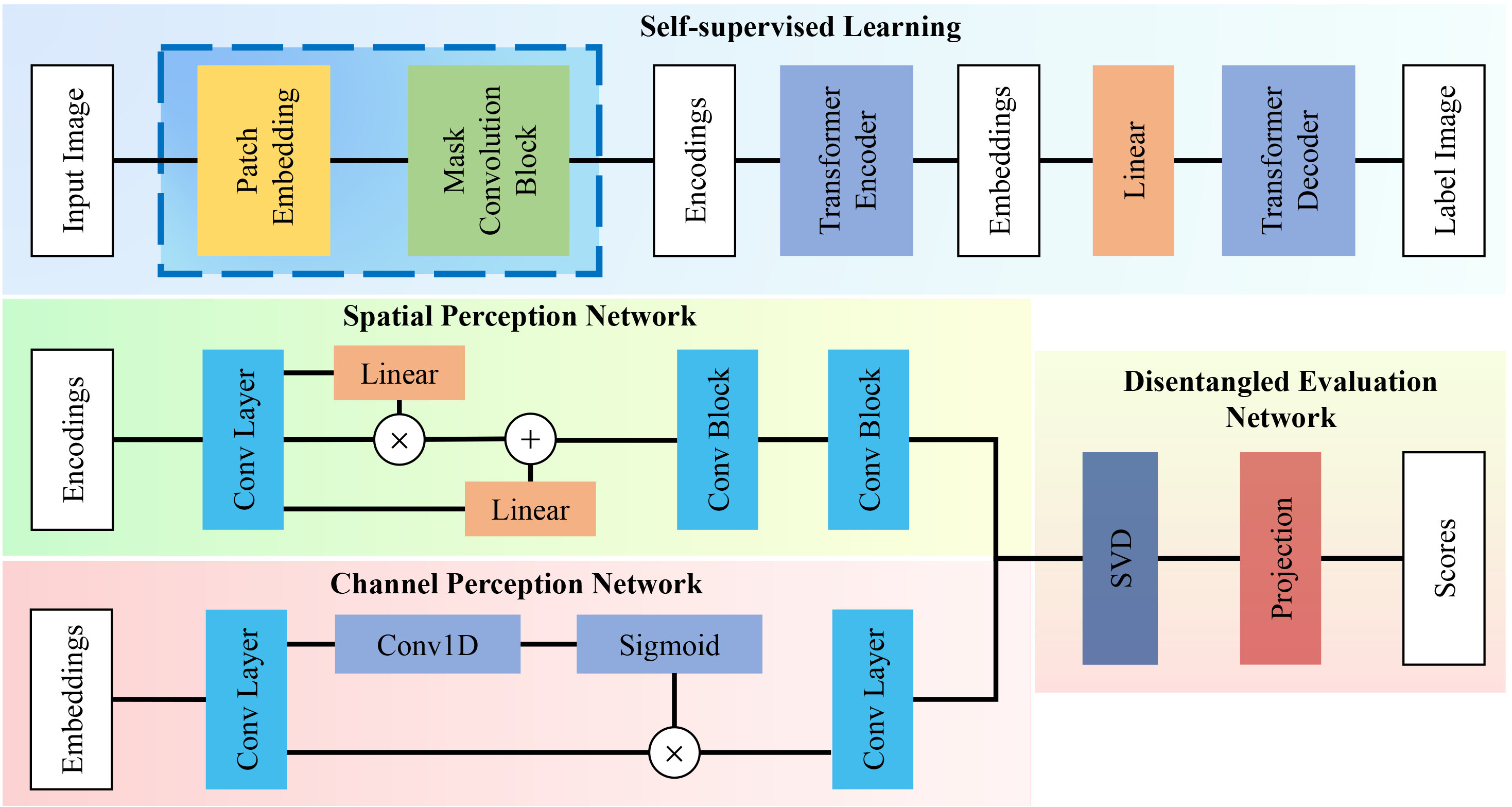}
	\caption{Our network consists of four parts: self-supervised learning, a spatial perception network, a channel perception network and a disentangled evaluation network.}
	\label{Network}
\end{figure*}

\section{Model}
\subsection{Overview}
The aesthetics assessment of children's paintings presents several challenges. First, insufficient labeled data may limit network learning capacity, inducing prediction biases. Second, the model needs to consider not only the aesthetic features of the images, but also the semantic and emotional content of the paintings. Third, the eight attributes of children's paintings present a complex mapping problem that can make model training unstable. To address these challenges, we propose a four-module network architecture (Figure~\ref{Network}). Through careful design and integration, we aim to enhance network performance on AACP.

\subsection{Self-supervised Learning Network}
To address the issue of limited labeled data, we employ a self-supervised learning strategy. As shown in Figure~\ref{Network}, we utilize a structure similar to ConvMAE~\cite{DBLP:journals/corr/abs-2205-03892}. During the training phase, generated images of children's paintings are used to train the model. The masking ratio is set to 0.75. After training for 1500 epochs, our model achieves 86\% accuracy on the test set. In the fine-tuning phase, we train the network on real children's paintings without masking. After fine-tuning for 20 epochs, the model obtains 93\% accuracy on real children's paintings. The latent codes from this module will be used in the prediction model, which includes the perception modules and the evaluation module and is described as follows:
\begin{equation}
	S = F (S_{e}(x, \theta_{s}), \theta_{F}),
\end{equation}
where $S$ represents the predicted aesthetic score, $F$ indicates the prediction model, $S_{e}$ means the encoder in self-supervised model, and $x$ represents the input image.

\subsection{Spatial Perception Network}
The spatial perception module aims to preserve spatial information of children's paintings, which is a crucial factor reflecting a child's psychological state that may be lost during encoding, leading to inaccurate aesthetic scores. To enhance scoring precision, we implement a spatial perception module which, as depicted in Figure~\ref{Network}, fuses the latent encodings learned from the self-supervised module into each convolution layer. The process can be expressed as:
\begin{equation}
	S = \omega_{\sigma}(e) \frac{F_{i} - \mu(F_{i})}{\sigma(F_{i})} + \omega_{\mu}(e),
\end{equation}
where $F_{i}$ denotes an intermediate feature map from each convolution layer. $e$ is the latent encodings from the self-supervised module, and $\omega_{\sigma}$ and $\omega_{\mu}$ are the learnable parameters for the standard deviation and mean, respectively. The function $\mu(F_{i})$ and $\sigma(F_{i})$ compute the channel-wise mean and standard deviation of $F_{i}$, respectively.

While our structure is similar to EQGAN-SA~\cite{DBLP:conf/cvpr/WangYXSLZ22}, their goal is to maintain spatial positions, requiring Gaussian distribution sampling. In contrast,  our model incorporates features obtained from self-supervised learning into the convolution module, ensuring that spatial information is not lost and effectively captures the intrinsic meanings of children's paintings, improving the performance of the network on AACP.

\subsection{Channel Perception Network}
The channel perception network is designed to extract channel information. In the aesthetics assessment of children's paintings, the channel information affects the accuracy of the aesthetic score. Some methods such as SE-NET~\cite{DBLP:conf/cvpr/HuSS18} and DAN~\cite{DBLP:conf/cvpr/FuLT0BFL19} use attention mechanism to learn the channel information, but they reduce the number of channels to reduce the computational effort. Therefore, to capture the channel information, we use a learning-based cross-channel module. 

First, the latent embeddings $x^{N \times L \times E}$ is transposed to $z^{N \times (LE)}$. Then, the output $x_{c}$ of our perception can be described as:
\begin{equation}
	x_{c} = \sigma (W^{k} (z)) \cdot x,
\end{equation}
where $W^{k}$ indicates a $1\mathit{D}$ convolution layer with kernel size $=$ k. In our experiments, we set k = 3 and padding $=$ 1, which are common parameter settings for obtaining weights.

\subsection{Disentangled Evaluation Network}
The evaluation module is designed to map spatial features and channel features to aesthetic scores. However, manually labeled aesthetic scores often contain small errors on each attribute. Previous methods that directly learn the mapping through fully connected layers result in direct interactions between different attributes, leading to lower accuracy. To address this issue, we adopt feature decoupling techniques in the model. There are various ways of feature decoupling, such as Principal Component Analysis (PCA)~\cite{DBLP:conf/cvpr/KeS04} and Singular Value Decomposition (SVD)~\cite{DBLP:journals/tsp/AharonEB06}. PCA treats the features as a high-dimensional vector and finds the correlation between feature maps by computing the covariance matrix and performing eigenvalue decomposition. This results in a disentangled representation. Similarly, SVD decomposes the features into matrices and retains the largest singular values to obtain a disentangled representation. 

Due to the non-square nature of the acquired feature maps in our study, SVD was employed as an alternative to PCA, which is unable to handle non-square data directly. We fuse spatial and channel features and apply SVD to project the results onto the eight aesthetic attributes. 

\begin{figure*}[t]
	\centering
	\includegraphics[width=1\linewidth]{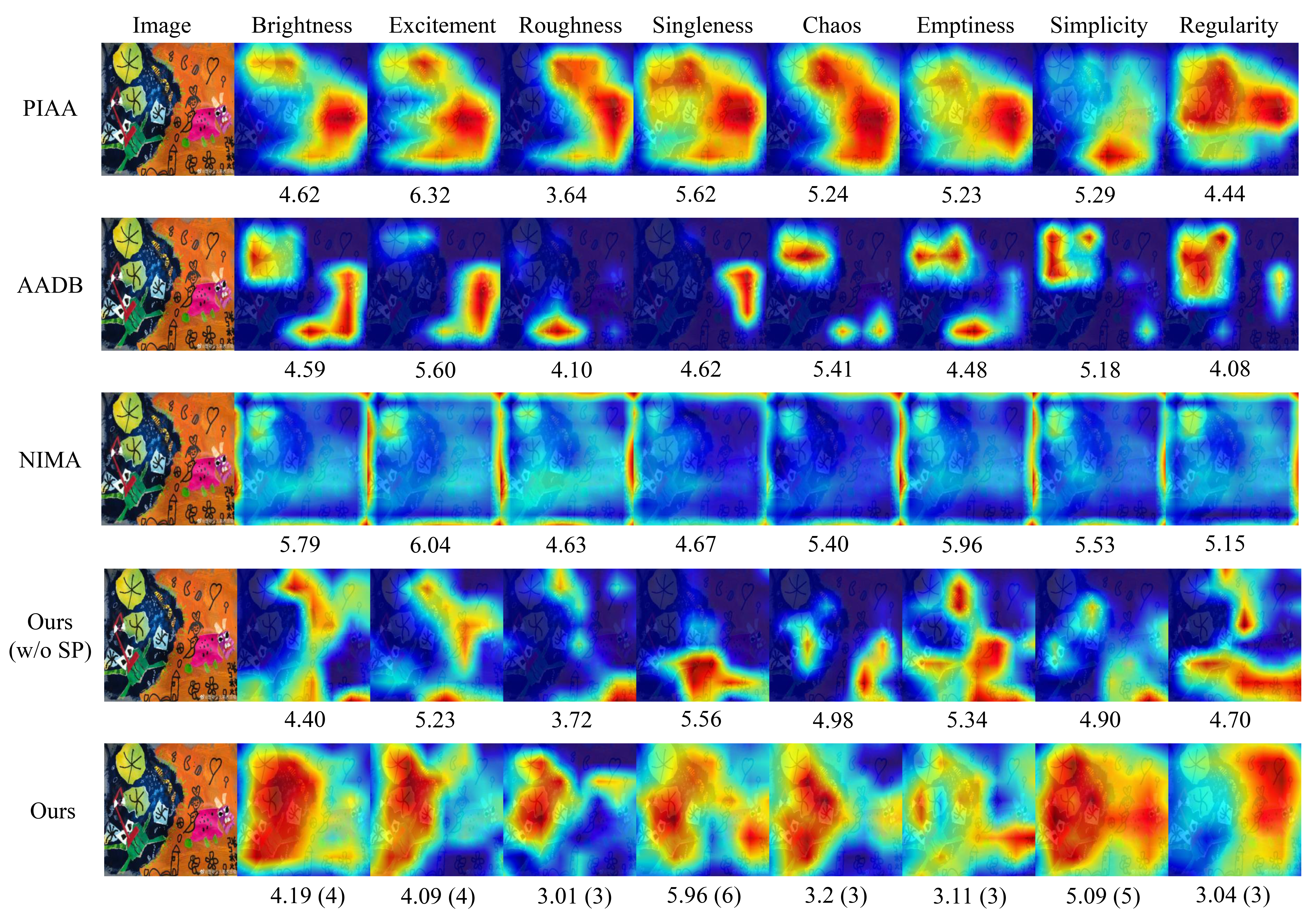}
	\caption{Spatial visual attention at intermediate layers, visualized by SmoothGradCAM. A strong contribution to the final score is indicated by a bright color. The last line shows the result of our method (ground truth).}
	\label{eva1}
\end{figure*}

\subsection{Training Details}
Our model is trained on an NVIDIA RTX 3090 using PyTorch, and takes 256 $\times$ 256 fixed images as the input. The training process is divided into two steps. In the first step, we use the default configurations to obtain the parameters of the self-supervised representation model. In the second step, we train the remaining modules without data augmentation. The Adam algorithm is used to optimize the model, and the Mean Squared Error (MSE) is used as the loss function. The model converges after 400 epochs of training with a learning rate of $1 \times 10^{-4}$ and a batch size of $64$.

\section{Evaluation}

\subsection{Qualitative Evaluation}
Figure~\ref{eva1} illustrates our proposed method's performance evaluation on spatial features, leveraging SmoothGradCam~\cite{DBLP:journals/corr/abs-1908-01224} for feature extraction.
We compare the feature maps of the last convolution layer in the spatial perception module. Our method identifies more activation regions than both NIMA and AADB, and when compared to PIAA, it shows varied activation region differences across attributes, showing our method's accuracy in feature extraction and disentanglement. An ablation study, removing the spatial perception module, confirmed its effectiveness by revealing a significant decrease in activation regions, attesting to the module's capability in aesthetic feature perception.

\subsection{Quantitative Evaluation}
Spearman's rank correlation coefficient (SRCC)~\cite{DBLP:journals/tip/TalebiM18} and the linear correlation coefficient (LCC)~\cite{DBLP:journals/tip/TalebiM18} are commonly used metrics for quantifying the results of evaluation models. The Earth Mover's Distance (EMD)~\cite{DBLP:journals/tip/TalebiM18} and MSE are two commonly utilized methods for calculating errors. In our quantitative experiments, we primarily employ these four metrics for comparison with other IAA models. All models were trained on the children's painting dataset and utilized the recommended parameter settings. As shown in Table~\ref{comp}, our method outperformed in most metrics, except EMD, where A LAMP was
slightly better. This indicates that incorporating spatial and channel information and utilizing feature disentanglement in the analysis of children's paintings fundamentally improves the perception of aesthetics in such images.
\begin{table}[t]
	\centering
	\caption{Comparison of 5 state-of-the-art IAA models on the CP dataset. For all models with publicly available codes, we use the recommended parameter settings.}
	\begin{tabular}{lccccc}
		\hline
		Metric & SRCC~$\uparrow$ & LCC~$\uparrow$  &  EMD~$\downarrow$ & MSE~$\downarrow$  \\ \hline
		NIMA      & 0.17  & 0.17  & 1.38  & 0.55  \\
		AADB      & 0.21  & 0.23  & 0.59  & 0.49  \\
		MLSP      & 0.36  & 0.39  & 0.62  & 0.42     \\ 
		A\_LAMP   & 0.05  & 0.04  & \textbf{0.14}  & 0.46     \\ 
		PIAA      & 0.27  & 0.30   & 0.15  & 0.45       \\\hline
		\textbf{Ours}      &\textbf{0.61}  & \textbf{0.65}  & 0.38    & \textbf{0.08}     \\\hline
	\end{tabular}
	\label{comp}
\end{table}
\subsection{User Study}
We conducted user studies to evaluate our dataset and model in terms of three aspects: \emph{dataset expansion}, \emph{painting annotation}, and \emph{aesthetics assessment}. We invited 15 participants, consisting of 10 non-experts and 5 experts. 

\textbf{Dataset Expansion}~~To assess the dissimilarity between generated and real children's paintings, we conducted an experiment where participants first viewed 20 real children's paintings. Subsequently, they were presented with 15 pairs of images, each pair containing a generated and a real painting, and asked to identify the real one. They were allowed to select both if both appeared real. Additionally, participants rated 15 generated paintings on a scale of 0 to 5, with 0 indicating significant deviation and 5 denoting strong similarity to real paintings.

\begin{figure}[H]
	\centering
	\includegraphics[width=1\linewidth]{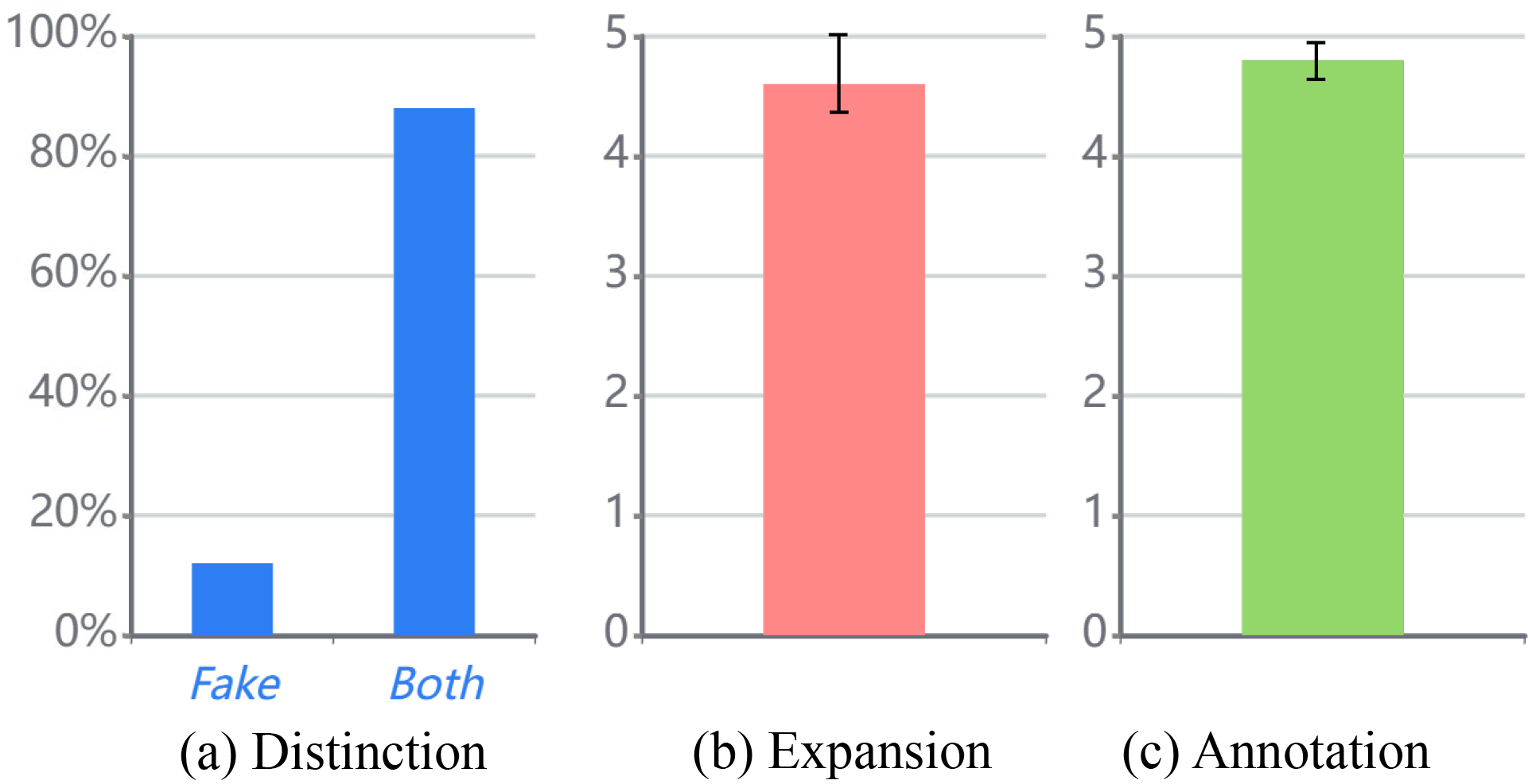}
	\caption{Statistic results of the user study. }
	\label{user}
\end{figure}

As shown in Figure~\ref{user} (a), 12\% of participants identified the generated paintings as real, while 88\% selected both.  Additionally, Figure~\ref{user} (b) indicates the dataset expansion yielded an average score of 4.6, suggesting that users found it challenging to distinguish between real and generated images. Thus, we conclude that the generated paintings, due to their similarity to real ones, are appropriate for dataset augmentation.

\textbf{Painting Annotation}~~To verify the validity and accuracy of our dataset annotations, we randomly selected 10 images. Participants were asked to observe and rate the reasonableness of scores for each attribute on a scale of 0 to 5. A rating of 0 indicates that the annotations are unreasonable, while a rating of 5 indicates that they are reasonable. As depicted in Figure~\ref{user} (c), the average rating for our painting annotations was found to be 4.8, indicating that our annotations are consistent with human perception and reflect a high level of accuracy. Therefore, our dataset is reliable.

\begin{table}[H]
	\centering
	\caption{Results of user study comparing the performance of various methods on AACP.}
	\begin{tabular}{lcc}
		\hline
		Method  & Mean~$\uparrow$ & Std~$\downarrow$ \\\hline
		
		MLSP    & 9.4         & 2.1        \\
		A\_LAMP & 9.5         & 1.6        \\
		PIAA    & 9.2         & 0.9        \\
		AADB    & 8.7         & 1.3        \\
		NIMA    & 8.9         & 1.1        \\\hline
		
		\textbf{Ours}    & \textbf{9.7}        & \textbf{0.8}        \\\hline
		
	\end{tabular}
	\label{user study}
\end{table}

\textbf{Aesthetics Assessment}~~We conducted a user study involving 15 participants to validate our method, comparing it with five other well-known methods. Participants evaluated the aesthetic assessment results of 25 randomly selected children's paintings, providing ratings on a scale of 1 to 10, with 10 indicating the most reasonable aesthetic score. Considering the subjective nature of aesthetics, diverse ratings provide a general consensus on the aesthetic qualities of the paintings.

We collect approximately 1200 judgments. The mean and standard error of these assessments are presented in Table~\ref{user study}. Our method obtains the highest mean score, indicating effective feature extraction from children's paintings can improve assessment accuracy. Moreover, our method achieves the lowest standard error, showing stability due to its disentangled nature. 

\subsection{Ablation Study}
In the ablation study, we evaluate the effectiveness of each component of our model. Our results, as presented in Table~\ref{A1}, indicate a significant decrease in performance when the self-supervised learning component is removed. This finding confirms the importance of including self-supervised learning in our model. We also found that the disentangled network component plays a crucial role, as evidenced by a significant decrease in the MSE metric when it is removed.

Additionally, our experiments demonstrated that the spatial and channel perception modules contribute to the overall performance of the model. Without these components, there is a slight decrease in all metrics. Therefore, we conclude that all four parts of our model are necessary to enhance the assessment ability of our method.
\begin{table}[]
	\centering
	\caption{Ablation studies of our network on AACP dataset. Each score is calculated as the average of the scores of the 8 attributes.}
	\begin{tabular}{lccc}
		\hline
		Type (W/O)          & SRCC~$\uparrow$ & LCC~$\uparrow$ & MSE~$\downarrow$          \\\hline
		SSL       & 0.16 & 0.20 & 0.12          \\	
		DE        & 0.49 & 0.48 & 0.14          \\
		CP        & 0.55 & 0.56 & 0.11          \\
		SP        & 0.41 & 0.42 & 0.13          \\\hline
		\textbf{Full Model} & \textbf{0.61} &\textbf{ 0.65} & \textbf{0.08} \\ \hline
	\end{tabular}	
	\label{A1}
\end{table}

\section{Conclusion}
In this paper, we first construct a dataset consisting of 20k unlabeled generated children's paintings and 1.2k manually labeled real children's paintings with eight attributes. Then, we design a model that includes a self-supervised learning module, a spatial perception module, a channel perception module and a disentangled evaluation module. Both quantitative experiments and user studies show that our method achieves SOTA performance on the aesthetics assessment of children's paintings. We also conduct ablation studies to investigate the impact of each module.

In the future, we will explore the aesthetic attributes of children's paintings more comprehensively and from varied dimensions to better understand children's aesthetic standards and creative expression. Furthermore, we plan to investigate the impact of environmental factors on AACP to uncover how various environmental factors may influence the development and expression of children's aesthetic sensibilities, providing valuable insights into how to cultivate creativity in young children.

\section{Acknowledgments}
This work was supported by the NSSFC under Grant 22ZD05 and the Shanghai Committee of Science and Technology, China (Grant No. 22511104600).

\bibliography{aaai24}

\end{document}